\documentclass[sigconf]{acmart}

\renewcommand\footnotetextcopyrightpermission[1]{}

\usepackage{amsthm}
\usepackage{booktabs}
\graphicspath{{figures/}}
\usepackage{microtype}
\usepackage{multirow}
\colorlet{Cblue}{blue!70!black}    
\colorlet{Corange}{orange!85!black}
\colorlet{Cgreen}{teal!65!black}   
\colorlet{Cdark}{gray!85!black}    
\colorlet{Cviolet}{blue!45!violet} 

\usepackage{tikz}
\usepackage{pgfplots}
\pgfplotsset{compat=1.18}
\usetikzlibrary{arrows.meta,positioning,shapes.geometric,fit,backgrounds,calc,decorations.pathreplacing}
\usepackage{enumitem}

\copyrightyear{2026}
\setcopyright{none}
\acmConference{Preprint}{2026}{}
\acmBooktitle{}
\acmDOI{}
\acmISBN{}

\fancypagestyle{plain}{\fancyhf{}\fancyfoot[C]{\thepage}}
\fancypagestyle{empty}{\fancyhf{}\fancyfoot[C]{\thepage}}
\fancypagestyle{fancy}{\fancyhf{}\fancyfoot[C]{\thepage}}
\AtBeginDocument{\pagestyle{plain}}

\begin{document}

\title{Gradient-Free Continual Learning in Spiking Neural
  Networks via Inter-Spike Interval Regularization}

\author{Samrendra Roy}
\email{roysam@illinois.edu}
\affiliation{%
  \institution{University of Illinois Urbana-Champaign}
  \department{Department of Nuclear, Plasma \& Radiological Engineering}
  \city{Urbana}
  \state{IL}
  \country{USA}
}

\author{Kazuma Kobayashi}
\affiliation{%
  \institution{University of Illinois Urbana-Champaign}
  \department{Department of Nuclear, Plasma \& Radiological Engineering}
  \city{Urbana}
  \state{IL}
  \country{USA}
}

\author{Souvik Chakraborty}
\affiliation{%
  \institution{Indian Institute of Technology Delhi}
  \department{Department of Applied Mechanics}
  \city{New Delhi}
  \country{India}
}

\author{Sajedul Talukder}
\affiliation{%
  \institution{University of Texas at El Paso}
  \department{Department of Computer Science}
  \city{El Paso}
  \state{TX}
  \country{USA}
}

\author{Syed Bahauddin Alam}
\affiliation{%
  \institution{University of Illinois Urbana-Champaign}
  \department{Department of Nuclear, Plasma \& Radiological Engineering}
  \city{Urbana}
  \state{IL}
  \country{USA}
}
\affiliation{%
  \institution{National Center for Supercomputing Applications}
  \city{Urbana}
  \state{IL}
  \country{USA}
}


\begin{abstract}
Continual learning, the ability to acquire new tasks sequentially without
forgetting prior knowledge, is essential for deploying neural networks in
dynamic real-world environments, from nuclear digital twin monitoring to
grid-edge fault detection. Existing synaptic importance methods, such as
Elastic Weight Consolidation (EWC) and Synaptic Intelligence (SI), rely on
gradient computation, making them incompatible with neuromorphic
hardware that lacks backpropagation support. We propose \textbf{ISI-CV}, the
first gradient-free synaptic importance metric for SNN continual learning,
derived from the Coefficient of Variation (CV) of Inter-Spike Intervals
(ISIs). Neurons that fire regularly (low CV) encode
stable, task-relevant features and are protected from overwriting; neurons
with irregular firing are permitted to adapt freely. ISI-CV requires only
spike time counters and integer arithmetic, all of which are native to
every neuromorphic chip. We evaluate on four benchmarks of increasing
difficulty: Split-MNIST, Permuted-MNIST, Split-Fashion\-MNIST, and
Split-N-MNIST using real Dynamic Vision Sensor (DVS) event data.
Across three seeds, ISI-CV achieves zero forgetting
(AF\,$= 0.000 \pm 0.000$) on Split-MNIST and Split-FashionMNIST,
near-zero forgetting on Permuted-MNIST (AF\,$= 0.001 \pm 0.000$),
and the highest accuracy with the lowest forgetting on real neuromorphic
DVS data (AA\,$= 0.820 \pm 0.012$, AF\,$= 0.221 \pm 0.014$). On
N-MNIST, gradient-based methods produce unreliable importance estimates
and perform worse than no regularization; ISI-CV avoids this failure by
design.
\end{abstract}


\maketitle
\pagestyle{plain}
\thispagestyle{plain}

\section{Introduction}
\label{sec:intro}

The ability to accumulate knowledge from a continuous stream of tasks without
catastrophically overwriting previously learned representations is a
fundamental requirement for intelligent systems deployed at the edge,
including nuclear digital twins that must adapt across fuel cycles and
grid-edge monitors tracking evolving fault signatures.
Catastrophic forgetting~\cite{mccloskey1989} remains the primary obstacle:
when a network is trained sequentially, gradient-descent updates that minimize
loss on the new task inadvertently destroy weight configurations that encoded
prior tasks. Regularization-based methods such as Elastic Weight
Consolidation (EWC)~\cite{kirkpatrick2017} and Synaptic
Intelligence (SI)~\cite{zenke2017} address this in standard artificial neural
networks (ANNs) by identifying and protecting important parameters, but both
require backpropagation to estimate importance. This creates an incompatibility
with the primary deployment target for energy-efficient edge AI:
\emph{neuromorphic hardware}. Platforms such as Intel Loihi~\cite{davies2018},
IBM TrueNorth~\cite{merolla2014}, and SpiNNaker~\cite{furber2014} perform
computation through asynchronous binary spike events without floating-point
arithmetic or gradient propagation. Sandia National Laboratories has
recently partnered with SpiNNcloud to investigate neuromorphic computing
for nuclear missions using
SpiNNaker2~\cite{sandia2024spinncloud}, reflecting growing demand for
spike-based computation in critical infrastructure.
Consequently, no existing continual learning importance method can run on-chip.

Spiking Neural Networks (SNNs) are the natural computational model for
neuromorphic hardware. Beyond architectural compatibility, SNNs produce a
temporal signal that ANNs do not: the precise \emph{timing} of individual
spikes. In neuroscience, the Coefficient of Variation (CV) of Inter-Spike
Intervals (ISIs) is a well-established measure of neuronal firing regularity
that distinguishes neurons encoding stable features (low CV, clock-like firing)
from those with noisy or uncommitted responses (high CV, irregular
firing)~\cite{softky1993}.

We exploit this biological insight for continual learning. After training on
task $k$, we compute the ISI-CV of each hidden neuron from a brief inference-only
pass over a representative batch of task-$k$ data. Neurons with low CV are inferred to encode task-critical
representations and are protected more strongly when the network learns
task $k{+}1$. The entire importance estimation pipeline requires only spike
counters, subtraction, and mean/variance arithmetic; all are available on
every neuromorphic chip without modification.

\paragraph{Contributions}
\begin{enumerate}[leftmargin=1.5em,topsep=2pt,itemsep=1pt]
  \item ISI-CV: the first gradient-free synaptic importance metric for
    SNN continual learning, derived entirely from spike timing statistics and
    requiring no backpropagation. (Training itself uses standard surrogate
    gradients; the gradient-free property applies to the importance
    estimation that runs between tasks and must execute on-chip.)
  \item Empirical demonstration across four benchmarks (three seeds each),
    including real DVS neuromorphic sensor data, that ISI-CV achieves zero
    or near-zero catastrophic forgetting while matching or outperforming
    gradient-based baselines on accuracy.
  \item A documented failure mode: gradient-based importance (EWC, SI)
    produces unreliable estimates on harder SNN benchmarks, performing
    worse than no regularization on N-MNIST; ISI-CV avoids this by
    operating on spike timing rather than approximate gradients.
\end{enumerate}

\paragraph{Motivating application}
While ISI-CV is validated on standard benchmarks, we
are motivated by a concrete deployment gap. Digital
twin frameworks for nuclear systems require edge-deployed models that
adapt to sequential operational changes without forgetting prior
regimes~\cite{kobayashi2024deeponet}, and grid-edge monitors must
sequentially learn fault signatures as conditions evolve.
Neuromorphic processors are suited to these roles: ultra-low-power,
always-on, and under active investigation for radiation-tolerant
deployment~\cite{sandia2024spinncloud}. ISI-CV enables on-chip continual
learning for such deployments without architectural modification.

\section{Related Work}
\label{sec:related}

Parameter regularization methods protect important weights by penalizing
deviations from their task-$k$ values during task-$(k{+}1)$ training.
EWC~\cite{kirkpatrick2017} uses the diagonal Fisher Information Matrix as an
importance proxy, computed via squared gradient magnitudes after task training.
SI~\cite{zenke2017} accumulates an online path integral of
gradient\,$\times$\,weight-change products during training, attributing
importance to parameters that contributed most to loss reduction.
Both require access to backpropagation and are therefore restricted to
GPU-based training environments. A comprehensive survey of continual learning is provided
in~\cite{van2024continual}.

Within the SNN domain, early continual learning work explored training with local
spike-timing-dependent plasticity (STDP) rules and biologically
inspired synaptic importance methods~\cite{antonov2022}.
More recently, Bayesian continual learning via SNNs~\cite{jang2022} and
multi-timescale astrocyte-modulated plasticity~\cite{haas2025} have been
proposed. However, all existing methods either require gradient computation
(via surrogate gradients or backpropagation-through-time) or operate only in
unsupervised settings. To our knowledge, no prior work has proposed a
gradient-free importance metric intrinsic to spiking computation.

The ISI statistics that ISI-CV builds upon have a long history.
Softky and Koch~\cite{softky1993} established that ISI-CV encodes fundamental
properties of cortical computation. ISI regularity has subsequently been linked
to encoding reliability in primary visual cortex~\cite{gur1997}. In the SNN
literature, Kim et al.~\cite{kim2021} use short ISI activity as a proxy for
neuron saliency in gradient-free visual explanations (Spike Activation Maps).
We are the first to apply ISI-CV as a \emph{synaptic importance} metric for
protecting against catastrophic forgetting.

On the hardware side, the deployment of neuromorphic processors in safety-critical settings is
accelerating. Sandia National Laboratories has partnered with SpiNNcloud to
explore neuromorphic architectures for nuclear deterrence
missions~\cite{sandia2024spinncloud}, and SNNs have recently been
demonstrated to solve finite element PDE problems natively on neuromorphic
hardware~\cite{theilman2025}. Deep neural operators have enabled
real-time virtual sensing for nuclear reactor
monitoring~\cite{kobayashi2024deeponet,hossain2025virtual}. These
developments establish the hardware platform and application demand;
ISI-CV contributes the missing piece: a continual learning mechanism
that operates within the constraints of spike-based computation.

\section{Method: ISI-CV Regularization}
\label{sec:method}

\subsection{Multi-Head SNN Architecture}

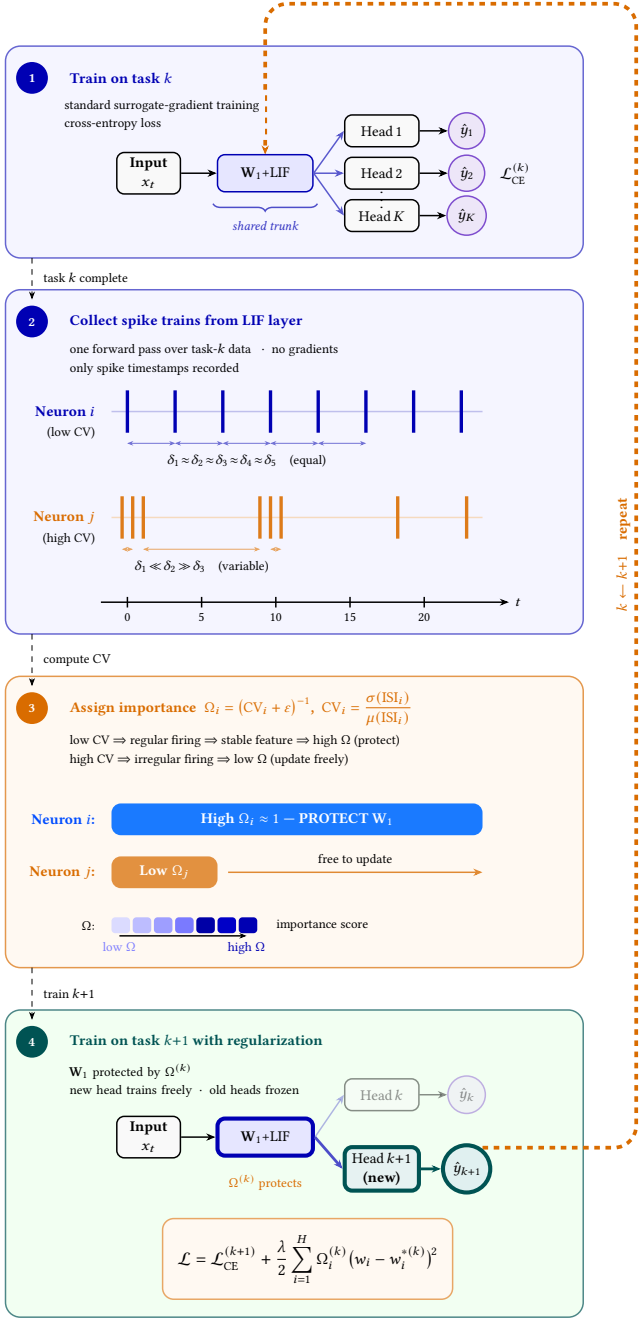
\begin{figure}[htbp]
\centering
\resizebox{\columnwidth}{!}{%
\begin{tikzpicture}[x=1cm,y=1cm,
  BLK/.style={rectangle,rounded corners=4pt,thick,
              minimum height=0.7cm,align=center,font=\small},
  ARCH/.style={BLK,draw=Cblue,fill=blue!10},
  HEAD/.style={BLK,draw=black,fill=gray!5,minimum height=0.6cm},
  OUTP/.style={circle,draw=Cviolet!70,fill=Cviolet!10,thick,
               minimum size=0.6cm,font=\small},
  IBOX/.style={BLK,draw=black,fill=gray!5},
  arr/.style={-{Stealth[length=5pt,width=4pt]},thick},
  sarr/.style={-{Stealth[length=4pt,width=3pt]},thin,black},
  darr/.style={arr,dashed,Corange},
]

\def\Lc{0.0} \def\Tx{0.6} \def\Rc{0.3}
\def\Nx{2.2} \def\Wx{4.4} \def\Hx{6.6}
\def\Yx{8.2} \def\BgR{10.4} \def\RX{11.4}

\def\Ya{9.5}

\begin{pgfonlayer}{background}
  \fill[blue!4,rounded corners=8pt] (-0.5,\Ya-3.4) rectangle (\BgR,\Ya+0.6);
  \draw[Cblue!60,rounded corners=8pt,thick] (-0.5,\Ya-3.4) rectangle (\BgR,\Ya+0.6);
\end{pgfonlayer}

\fill[Cblue] (\Lc,\Ya) circle (\Rc)
  node[white,font=\footnotesize\bfseries] {1};

\draw[darr,Corange,line width=2pt,rounded corners=12pt]
  (\Yx+0.3,-10.7) -- (\RX,-10.7) -- (\RX,\Ya+1.4) --
  (\Wx,\Ya+1.4) -- (4.4,\Ya-1.45);
\node[font=\small\bfseries,Corange,rotate=90,anchor=south]
  at (\RX,0.5) {$k \leftarrow k{+}1$\quad repeat};

\node[font=\small\bfseries,Cblue,anchor=west] at (\Tx,\Ya)
  {Train on task $k$};
\node[font=\footnotesize,black,anchor=north west,align=left,
      fill=blue!4,inner sep=3pt]
  at (\Tx-0.1,\Ya-0.3) {%
  standard surrogate-gradient training\\[2pt]cross-entropy loss};

\node[IBOX,minimum width=1.2cm] (IN1) at (\Nx,\Ya-1.8)
  {\textbf{Input}\\$x_t$};
\node[ARCH,minimum width=1.8cm] (W1)  at (\Wx,\Ya-1.8)
  {$\mathbf{W}_1$+LIF};
\node[HEAD,minimum width=1.4cm] (HA)  at (\Hx,\Ya-1.0) {Head\,$1$};
\node[HEAD,minimum width=1.4cm] (HB)  at (\Hx,\Ya-1.8) {Head\,$2$};
\node[HEAD,minimum width=1.4cm] (HC)  at (\Hx,\Ya-2.6) {Head\,$K$};
\node[font=\small,black]           at (\Hx,\Ya-2.2) {$\vdots$};
\node[OUTP] (Y1) at (\Yx,\Ya-1.0) {$\hat y_1$};
\node[OUTP] (Y2) at (\Yx,\Ya-1.8) {$\hat y_2$};
\node[OUTP] (YK) at (\Yx,\Ya-2.6) {$\hat y_K$};

\draw[arr,black]  (IN1.east) -- (W1.west);
\foreach \h in {HA,HB,HC} \draw[arr,Cblue!65] (W1.east) -- (\h.west);
\draw[arr,black]  (HA.east)  -- (Y1.west);
\draw[arr,black]  (HB.east)  -- (Y2.west);
\draw[arr,black]  (HC.east)  -- (YK.west);

\draw[decorate,decoration={brace,amplitude=4pt,mirror},Cblue!70,thick]
  ([xshift=-2pt,yshift=-6pt]W1.south west) --
  ([xshift=2pt,yshift=-6pt]W1.south east)
  node[midway,below=6pt,font=\footnotesize\itshape,Cblue!80]
  {shared trunk};

\node[font=\small,black,anchor=west] at (\Yx+0.5,\Ya-1.8)
  {$\mathcal{L}_{\mathrm{CE}}^{(k)}$};

\draw[sarr,dashed] (\Lc,\Ya-3.4) -- (\Lc,5.32)
  node[midway,right=3pt,font=\footnotesize,black]{task $k$ complete};

\def\Yb{4.9}
\def\SL{1.5} \def\SR{8.5} \def\SH{0.4}

\begin{pgfonlayer}{background}
  \fill[blue!4,rounded corners=8pt] (-0.5,\Yb-5.9) rectangle (\BgR,\Yb+0.6);
  \draw[Cblue!60,rounded corners=8pt,thick] (-0.5,\Yb-5.9) rectangle (\BgR,\Yb+0.6);
\end{pgfonlayer}

\fill[Cblue] (\Lc,\Yb) circle (\Rc)
  node[white,font=\footnotesize\bfseries] {2};
\node[font=\small\bfseries,Cblue,anchor=west] at (\Tx,\Yb)
  {Collect spike trains from LIF layer};
\node[font=\footnotesize,black,anchor=north west,align=left]
  at (\Tx,\Yb-0.3) {%
  one forward pass over task-$k$ data
  \enspace$\cdot$\enspace no gradients\\[2pt]%
  only spike timestamps recorded};

\def\Yi{\Yb-1.7}
\node[font=\small\bfseries,Cblue,anchor=east]  at (\SL-0.2,\Yi)     {Neuron $i$};
\node[font=\footnotesize,black,anchor=east]    at (\SL-0.2,\Yi-0.4) {(low CV)};
\draw[Cblue!25,thick] (\SL,\Yi) -- (\SR,\Yi);
\foreach \sx in {1.8,2.7,3.6,4.5,5.4,6.3,7.2,8.1}
  \draw[Cblue,line width=1.8pt] (\sx,\Yi-\SH) -- (\sx,\Yi+\SH);
\foreach \xa/\xb in {1.8/2.7,2.7/3.6,3.6/4.5,4.5/5.4,5.4/6.3}
  \draw[Cblue!55,thin,{Stealth[length=3pt]}-{Stealth[length=3pt]}]
    (\xa,\Yi-0.6) -- (\xb,\Yi-0.6);
\node[font=\footnotesize,black,anchor=north] at (4.05,\Yi-0.7)
  {$\delta_1 \!\approx\! \delta_2 \!\approx\! \delta_3 \!\approx\! \delta_4 \!\approx\! \delta_5$\quad(equal)};

\def\Yj{\Yb-3.7}
\node[font=\small\bfseries,Corange,anchor=east]  at (\SL-0.2,\Yj)     {Neuron $j$};
\node[font=\footnotesize,black,anchor=east]      at (\SL-0.2,\Yj-0.4) {(high CV)};
\draw[Corange!25,thick] (\SL,\Yj) -- (\SR,\Yj);
\foreach \sx in {1.7,1.9,2.1,4.3,4.5,4.7,6.9,8.2}
  \draw[Corange!90,line width=1.8pt] (\sx,\Yj-\SH) -- (\sx,\Yj+\SH);
\draw[Corange!60,thin,{Stealth[length=3pt]}-{Stealth[length=3pt]}]
  (1.7,\Yj-0.6) -- (1.9,\Yj-0.6);
\draw[Corange!60,thin,{Stealth[length=3pt]}-{Stealth[length=3pt]}]
  (2.1,\Yj-0.6) -- (4.3,\Yj-0.6);
\draw[Corange!60,thin,{Stealth[length=3pt]}-{Stealth[length=3pt]}]
  (4.5,\Yj-0.6) -- (4.7,\Yj-0.6);
\node[font=\footnotesize,black,anchor=north] at (3.2,\Yj-0.7)
  {$\delta_1 \!\ll\! \delta_2 \!\gg\! \delta_3$\quad(variable)};

\def\Yax{\Yb-5.3}
\draw[-{Stealth[length=5pt]},black,thick]
  (\SL-0.2,\Yax) -- (\SR+0.5,\Yax)
  node[right,font=\footnotesize,black]{$t$};
\foreach \x/\l in {1.8/0,3.2/5,4.6/10,6.0/15,7.4/20}{
  \draw[black,thick] (\x,\Yax+0.1) -- (\x,\Yax-0.1);
  \node[font=\footnotesize,black,anchor=north] at (\x,\Yax-0.1) {\l};
}

\draw[sarr,dashed] (\Lc,\Yb-5.9) -- (\Lc,-1.98)
  node[midway,right=3pt,font=\footnotesize,black]{compute $\mathrm{CV}$};

\def\Yc{-2.4}

\begin{pgfonlayer}{background}
  \fill[orange!5,rounded corners=8pt] (-0.5,\Yc-4.9) rectangle (\BgR,\Yc+0.6);
  \draw[Corange!70,rounded corners=8pt,thick] (-0.5,\Yc-4.9) rectangle (\BgR,\Yc+0.6);
\end{pgfonlayer}

\fill[Corange] (\Lc,\Yc) circle (\Rc)
  node[white,font=\footnotesize\bfseries] {3};
\node[font=\small\bfseries,Corange,anchor=west] at (\Tx,\Yc)
  {Assign importance\enspace
   $\Omega_i = \bigl(\mathrm{CV}_i + \varepsilon\bigr)^{-1}$,\enspace
   $\mathrm{CV}_i = \dfrac{\sigma(\mathrm{ISI}_i)}{\mu(\mathrm{ISI}_i)}$};
\node[font=\footnotesize,black,anchor=north west,align=left]
  at (\Tx,\Yc-0.4) {%
  low CV $\Rightarrow$ regular firing $\Rightarrow$ stable feature
  $\Rightarrow$ high $\Omega$ (protect)\\[2pt]%
  high CV $\Rightarrow$ irregular firing
  $\Rightarrow$ low $\Omega$ (update freely)};

\def\Yco{\Yc-2.1}
\node[font=\small\bfseries,blue!70!cyan,anchor=east] at (\SL-0.2,\Yco)
  {Neuron $i$:};
\fill[blue!58!cyan!90,rounded corners=5pt] (\SL,\Yco-0.3) rectangle (\SR,\Yco+0.3);
\node[font=\small\bfseries,white] at (5.0,\Yco)
  {High $\Omega_i \approx 1$\;—\;\textbf{PROTECT $\mathbf{W}_1$}};

\def\Yct{\Yc-3.1}
\node[font=\small\bfseries,Corange,anchor=east] at (\SL-0.2,\Yct)
  {Neuron $j$:};
\fill[Corange!75,rounded corners=5pt]
  (\SL,\Yct-0.3) rectangle (\SL+2.0,\Yct+0.3);
\node[font=\small\bfseries,white] at (\SL+1.0,\Yct) {Low $\Omega_j$};
\draw[arr,Corange!80,thick] (\SL+2.2,\Yct) --
  node[above,font=\footnotesize,black]{free to update} (\SR,\Yct);

\def\Ycs{\Yc-4.1}
\node[font=\footnotesize,black,anchor=east] at (\SL-0.2,\Ycs) {$\Omega$:};
\foreach \i/\col in {
  0/blue!14, 1/blue!26, 2/blue!38,
  3/blue!52, 4/blue!65!black, 5/blue!78!black, 6/Cblue}
  \fill[\col,rounded corners=2pt]
    (\SL+\i*0.4,\Ycs-0.15) rectangle (\SL+\i*0.4+0.35,\Ycs+0.15);
\node[font=\footnotesize,blue!50,anchor=north] at (\SL+0.15,\Ycs-0.2)
  {low $\Omega$};
\node[font=\footnotesize,Cblue,anchor=north]   at (\SL+2.55,\Ycs-0.2)
  {high $\Omega$};
\draw[-{Stealth[length=4pt]},black,thick]
  (\SL+0.15,\Ycs-0.2) -- (\SL+2.55,\Ycs-0.2);
\node[font=\footnotesize,black,anchor=west] at (\SL+3.0,\Ycs)
  {importance score};

\draw[sarr,dashed] (\Lc,\Yc-4.9) -- (\Lc,-8.28)
  node[midway,right=3pt,font=\footnotesize,black]{train $k{+}1$};

\def\Yd{-8.7}

\begin{pgfonlayer}{background}
  \fill[green!5,rounded corners=8pt] (-0.5,\Yd-5.2) rectangle (\BgR,\Yd+0.6);
  \draw[Cgreen!70,rounded corners=8pt,thick] (-0.5,\Yd-5.2) rectangle (\BgR,\Yd+0.6);
\end{pgfonlayer}

\fill[Cgreen] (\Lc,\Yd) circle (\Rc)
  node[white,font=\footnotesize\bfseries] {4};
\node[font=\small\bfseries,Cgreen,anchor=west] at (\Tx,\Yd)
  {Train on task $k{+}1$ with regularization};
\node[font=\footnotesize,black,anchor=north west,align=left,
      text width=5.0cm]
  at (\Tx,\Yd-0.3) {%
  $\mathbf{W}_1$ protected by $\Omega^{(k)}$\\[2pt]%
  new head trains freely\enspace$\cdot$\enspace old heads frozen};

\node[IBOX,minimum width=1.2cm] (IN4) at (\Nx,\Yd-1.8)
  {\textbf{Input}\\$x_t$};
\node[ARCH,minimum width=1.8cm,line width=2.2pt] (W14) at (\Wx,\Yd-1.8)
  {$\mathbf{W}_1$+LIF};
\node[HEAD,minimum width=1.4cm,opacity=0.45] (HA4) at (\Hx,\Yd-1.0)
  {Head\,$k$};
\node[HEAD,minimum width=1.4cm,
      draw=Cgreen,line width=2.2pt,fill=teal!10] (HB4) at (\Hx,\Yd-2.4)
  {Head\,$k{+}1$\\\textbf{(new)}};
\node[OUTP,opacity=0.45]                               (YA4) at (\Yx,\Yd-1.0)
  {$\hat y_k$};
\node[OUTP,draw=Cgreen,fill=teal!12,line width=2.2pt]  (YB4) at (\Yx,\Yd-2.4)
  {$\hat y_{k+1}$};

\draw[arr,black]                     (IN4.east) -- (W14.west);
\draw[arr,Cblue!65,opacity=0.45]        (W14.east) -- (HA4.west);
\draw[arr,Cblue!70,line width=1.5pt]    (W14.east) -- (HB4.west);
\draw[arr,black,opacity=0.45]        (HA4.east) -- (YA4.west);
\draw[arr,Cgreen,line width=1.5pt]      (HB4.east) -- (YB4.west);

\node[font=\footnotesize,Corange,anchor=north] at (\Wx,\Yd-2.4)
  {$\Omega^{(k)}$ protects};

\node[draw=Corange!60,fill=orange!5,rounded corners=6pt,thick,
      font=\normalsize,align=center,inner sep=8pt]
  at ({(\Nx+\Yx)/2},\Yd-4.1) {%
  $\mathcal{L} = \mathcal{L}_{\mathrm{CE}}^{(k+1)}
   + \dfrac{\lambda}{2}\displaystyle\sum_{i=1}^{H}
     \Omega_i^{(k)}\bigl(w_i - w_i^{*(k)}\bigr)^{\!2}$};

\end{tikzpicture}
}%
\Description{Four-step diagram showing ISI-CV continual learning cycle: train task k, collect spike trains, assign importance scores via CV, train task k+1 with regularization.}
\caption{%
  ISI-CV regularization as a four-step continual learning cycle.
  \textbf{(1)}~The SNN is trained on task~$k$ via surrogate-gradient
  descent. The shared trunk ($\mathbf{W}_1$+LIF) feeds $K$ per-task heads.
  \textbf{(2)}~After training, a gradient-free inference pass collects
  hidden spike trains. Neuron~$i$ (blue) fires regularly (equal ISIs,
  low CV); Neuron~$j$ (orange) fires in burst--gap bursts (variable
  ISIs, high CV).
  \textbf{(3)}~ISI-CV importance scores are assigned:
  $\Omega_i = (\mathrm{CV}_i{+}\varepsilon)^{-1}$.
  High-$\Omega$ neurons protect $\mathbf{W}_1$ (blue bar);
  low-$\Omega$ neurons update freely (orange bar).
  \textbf{(4)}~Task~$k{+}1$ trains with the regularized loss
  (Eq.~\ref{eq:loss}). The new head (teal) adapts freely while
  $\mathbf{W}_1$ is penalized by $\Omega^{(k)}$.
  The orange loop arrow marks the increment to task~$k{+}2$.%
}
\label{fig:architecture}
\end{figure}

We use a multi-head SNN with a shared feature trunk and per-task output
heads (Fig.~\ref{fig:architecture}). The shared trunk, comprising a fully-connected
layer $\mathbf{W}_1 \in \mathbb{R}^{H \times D}$ followed by a LIF neuron
layer of size $H$, is the target of regularization, as it encodes
task-shared representations. Each task $k$ is served by a dedicated head
$\mathbf{W}_2^{(k)} \in \mathbb{R}^{C \times H}$, where $C$ is the
number of classes per task.

The membrane potential of hidden LIF neuron $i$ at discrete timestep $t$ is:
\begin{equation}
  u_i^t = \tau \, u_i^{t-1} + \sum_j W_{ij}^{(1)} s_j^t - \theta \, s_i^{t-1}
  \label{eq:lif}
\end{equation}
where $\tau{=}2.0$ is the membrane time constant\footnote{In
SpikingJelly's LIF implementation, $\tau$ parameterizes exponential decay
as $v(t) = v(t{-}1)(1 - 1/\tau) + I(t)$; $\tau{=}2.0$ (the library default)
yields a decay factor of $0.5$ per timestep.}, $s_i^t \in \{0,1\}$ is the
binary spike output, and $\theta$ is the firing threshold. A spike is generated
when $u_i^t \geq \theta$, after which the membrane resets. The network is
trained end-to-end via surrogate gradient descent~\cite{neftci2019} using the
ATan surrogate function, implemented with SpikingJelly~\cite{fang2023}, with
task cross-entropy loss and $T{=}20$ timesteps per sample for MNIST-based
benchmarks ($T{=}10$ for N-MNIST).

For N-MNIST, a convolutional frontend processes each $[2,34,34]$ event frame
before the shared LIF trunk. It consists of two convolutional blocks
(Conv$_{2 \to 16}$, AvgPool, Conv$_{16 \to 32}$, AvgPool) followed by
a linear projection to 256 dimensions, enabling the model to handle
two-channel DVS spatial event data.

\subsection{ISI-CV Importance Estimation}

After completing training on task $k$, we perform a single inference pass
over a small subset of task-$k$ training data (no gradient computation)
to collect spike trains from the shared LIF layer. For each
neuron $i$, let $\mathcal{T}_i = \{t : s_i^t = 1\}$ be the set of spike times
across a batch of samples. The Inter-Spike Intervals are:
\begin{equation}
  \mathrm{ISI}_i = \bigl\{t_{j+1} - t_j \;\big|\;
    t_j,\,t_{j+1} \in \mathcal{T}_i,\; j = 1,\ldots,|\mathcal{T}_i|-1\bigr\}
  \label{eq:isi}
\end{equation}
The Coefficient of Variation and raw importance score are:
\begin{equation}
  \mathrm{CV}_i = \frac{\sigma(\mathrm{ISI}_i)}{\mu(\mathrm{ISI}_i) + \varepsilon},
  \qquad
  \tilde{\Omega}_i = \frac{1}{\mathrm{CV}_i + \varepsilon}
  \label{eq:cv}
\end{equation}
where $\varepsilon = 10^{-3}$ ensures numerical stability. Neurons with
$|\mathcal{T}_i| < 2$ (silent or near-silent) are assigned
$\mathrm{CV}_i = 2.0$, indicating low importance. The final importance
vector is obtained by clipping at the 95th percentile (to prevent outlier
neurons from dominating) and normalizing to $[0,1]$:
\begin{equation}
  \Omega_i = \frac{\min\!\bigl(\tilde{\Omega}_i,\,P_{95}\bigr)}{P_{95} + \varepsilon}
  \label{eq:norm}
\end{equation}

A neuron with low CV fires at near-constant intervals, indicating that the
input reliably and repeatedly drives it above threshold. This is a signature
of a neuron that has converged to a stable, task-specific feature detector.
A neuron with high CV fires sporadically, suggesting it has not settled
into a consistent encoding. Protecting low-CV neurons preserves the learned
task representation; permitting high-CV neurons to update maintains plasticity
for new tasks.

\subsection{Regularization Loss}

When training on task $k{+}1$, the total loss penalizes deviations of trunk
weights from their task-$k$ checkpoint $\mathbf{W}_1^{*(k)}$, weighted by
neuron importance:
\begin{equation}
  \mathcal{L} = \mathcal{L}_{\mathrm{CE}}^{(k+1)} +
  \frac{\lambda}{2} \sum_{i=1}^{H} \Omega_i
    \Biggl[\sum_{d=1}^{D}
      \bigl(W_{id}^{(1)} - W_{id}^{*(k)}\bigr)^2
    + \bigl(b_i - b_i^{*(k)}\bigr)^2
    \Biggr]
  \label{eq:loss}
\end{equation}
where $\lambda > 0$ controls the stability--plasticity tradeoff.
Per-task head weights $\mathbf{W}_2^{(k)}$ are always free to update.
Only the shared trunk weights are regularized, as they encode
features used by all tasks. ISI-CV assigns importance per neuron rather
than per synapse; this coarser granularity is a natural match for
neuromorphic hardware, where each neuron maps to a dedicated compute
core and a single importance value per core minimizes on-chip storage.

\subsection{Hardware Deployability}

The ISI-CV pipeline (Eqs.~\ref{eq:isi}--\ref{eq:norm}) reduces to:
(i)~a per-neuron spike timestamp counter, (ii)~integer subtraction to
compute ISIs, (iii)~fixed-point mean and variance over the ISI sequence.
These are elementary operations present on all major neuromorphic platforms
including Loihi~\cite{davies2018} and TrueNorth~\cite{merolla2014}.
The resulting importance vector $\Omega \in [0,1]^H$ is a compact weight
array (512 floats = 2\,kB) stored in SRAM between tasks.

In contrast, EWC requires a backward pass over task data to compute
$\mathbb{E}[(\nabla_w \mathcal{L})^2]$, and SI accumulates
$\nabla_w \mathcal{L} \cdot \Delta w$ throughout training. Both demand
floating-point gradient arithmetic not available on current neuromorphic
hardware. The computational cost difference is substantial: ISI-CV's
importance estimation is a single forward inference pass using only
integer spike counters, whereas EWC and SI require full
backpropagation with floating-point multiplications. On neuromorphic
chips operating at milliwatt-scale power budgets, this distinction
determines whether importance estimation can run on-chip at all.
ISI-CV is the first importance method that is natively executable in
this environment.

\begin{table}[t]
\caption{Split-MNIST (5 tasks, multi-head SNN, 3 seeds). Fixed $\lambda$:
  ISI-CV\,=\,500; EWC, SI\,=\,1000. \textbf{Bold} = best per column.}
\label{tab:split}
\centering\small
\setlength{\tabcolsep}{4pt}
\begin{tabular}{@{}lccc@{}}
\toprule
Method  & AA\,$\uparrow$ & BWT\,$\uparrow$ & AF\,$\downarrow$ \\
\midrule
No Reg  & $.982 {\scriptstyle\pm .004}$ & $-.018 {\scriptstyle\pm .006}$ & $.018 {\scriptstyle\pm .006}$ \\
EWC     & $.967 {\scriptstyle\pm .011}$ & $-.037 {\scriptstyle\pm .014}$ & $.037 {\scriptstyle\pm .014}$ \\
SI      & $\mathbf{.993} {\scriptstyle\pm .001}$ & $\mathbf{+.001} {\scriptstyle\pm .000}$ & $\mathbf{.000} {\scriptstyle\pm .000}$ \\
ISI-CV  & $.983 {\scriptstyle\pm .001}$ & $-.000 {\scriptstyle\pm .000}$ & $\mathbf{.000} {\scriptstyle\pm .000}$ \\
\bottomrule
\end{tabular}
\end{table}

\begin{table}[t]
\caption{Permuted-MNIST (5 tasks, multi-head SNN, 3 seeds).
  Fixed $\lambda$: ISI-CV\,=\,500; EWC, SI\,=\,1000.
  \textbf{Bold} = best per column.}
\label{tab:perm}
\centering\small
\setlength{\tabcolsep}{4pt}
\begin{tabular}{@{}lccc@{}}
\toprule
Method  & AA\,$\uparrow$ & BWT\,$\uparrow$ & AF\,$\downarrow$ \\
\midrule
No Reg  & $.929 {\scriptstyle\pm .005}$ & $-.052 {\scriptstyle\pm .005}$ & $.052 {\scriptstyle\pm .005}$ \\
EWC     & $\mathbf{.954} {\scriptstyle\pm .004}$ & $-.021 {\scriptstyle\pm .004}$ & $.021 {\scriptstyle\pm .004}$ \\
SI      & $.917 {\scriptstyle\pm .001}$ & $-.003 {\scriptstyle\pm .001}$ & $.003 {\scriptstyle\pm .001}$ \\
ISI-CV  & $.900 {\scriptstyle\pm .002}$ & $\mathbf{-.001} {\scriptstyle\pm .000}$ & $\mathbf{.001} {\scriptstyle\pm .000}$ \\
\bottomrule
\end{tabular}
\end{table}

\section{Experimental Setup}
\label{sec:experiments}

\paragraph{Benchmarks}
We evaluate on four sequential learning settings of increasing difficulty.
(1)~\emph{Split-MNIST} partitions the MNIST dataset~\cite{lecun1998} into
5 tasks of 2-class digit pairs: $\{0,1\},\ldots,\{8,9\}$.
(2)~\emph{Permuted-MNIST} applies five distinct random pixel permutations
to all 10 MNIST classes, yielding 5 tasks with the same output
labels but conflicting input statistics. Each permutation demands a
different spatial feature mapping from the shared trunk,
creating strong forgetting pressure on trunk weights.
(3)~\emph{Split-FashionMNIST} applies the same 5-task split to
FashionMNIST~\cite{xiao2017fashionmnist}, which contains visually more
complex clothing items instead of digits. The higher inter-class similarity
increases forgetting pressure relative to Split-MNIST.
(4)~\emph{Split-N-MNIST} draws five two-class tasks from
N-MNIST~\cite{orchard2015,lenz2021}, a neuromorphic dataset captured with a Dynamic
Vision Sensor (DVS) camera. Each sample is a stream of real spike events
$(x, y, t, \mathrm{pol})$ binned into $T{=}10$ temporal frames of shape
$[2, 34, 34]$. N-MNIST contains genuine neuromorphic sensor
dynamics, making it the most realistic test of ISI-CV in its
intended deployment context.

\paragraph{Architecture and training}
Split-MNIST, Permuted-MNIST, and Split-FashionMNIST use a
784\,$\to$\,512-LIF trunk with $T{=}20$ timesteps.
N-MNIST uses a convolutional frontend feeding a 256-LIF trunk
with $T{=}10$ timesteps (Sec.~\ref{sec:method}). All experiments use the
ATan surrogate gradient~\cite{neftci2019}, Adam optimizer
($\eta{=}10^{-3}$, 10 epochs per task), batch size 128, and FP16 mixed
precision training on an NVIDIA RTX 3060 (12\,GB). All results report
mean $\pm$ standard deviation across three independent seeds;
variance is consistently low for ISI-CV ($\sigma_{\mathrm{AF}} \leq 0.014$
across all benchmarks), indicating stable performance across
initializations.

\paragraph{Baselines}
We compare against: No Reg (sequential fine-tuning, no
regularization), EWC~\cite{kirkpatrick2017} with diagonal Fisher
importance, and SI~\cite{zenke2017} with online path-integral
importance. Both baselines require gradient computation and serve as
representative gradient-based methods from the continual learning literature.
We restrict comparison to regularization-based methods because they share
ISI-CV's mechanism (per-parameter importance weighting), enabling direct
comparison of importance estimation quality. Replay-based methods
(e.g., GEM~\cite{lopez2017}) require storing past task data, which conflicts
with memory-constrained neuromorphic deployment, and architecture-based
methods (e.g., PackNet, progressive networks) require dynamic graph
modification not supported on current neuromorphic chips.

\paragraph{Hyperparameter selection}
For Split-MNIST, Permuted-MNIST, and Split-FashionMNIST we use fixed
regularization strengths:
$\lambda_{\text{ISI}}{=}500$, $\lambda_{\text{EWC}}{=}\lambda_{\text{SI}}{=}1000$,
selected via a preliminary sweep on Split-MNIST. For N-MNIST,
Table~\ref{tab:nmnist} reports multi-seed results at fixed $\lambda$ values
($\lambda_{\text{ISI}}{=}\lambda_{\text{EWC}}{=}100$,
$\lambda_{\text{SI}}{=}1000$) to provide honest variance estimates at a
single operating point, while Fig.~\ref{fig:pareto}
shows the full single-seed sensitivity across
$\lambda \in \{10, 50, 100, 200, 500, 1000, 5000\}$.

\paragraph{Metrics}
Following~\cite{lopez2017}, we report three standard continual learning
metrics evaluated after training all $K$ tasks:
AA\,$\uparrow$ (Average Accuracy across all tasks),
BWT\,$\uparrow$ (Backward Transfer; negative values indicate
forgetting), and AF\,$\downarrow$ (Average Forgetting).

\begin{figure*}[htbp]
\centering
\includegraphics[width=\textwidth]{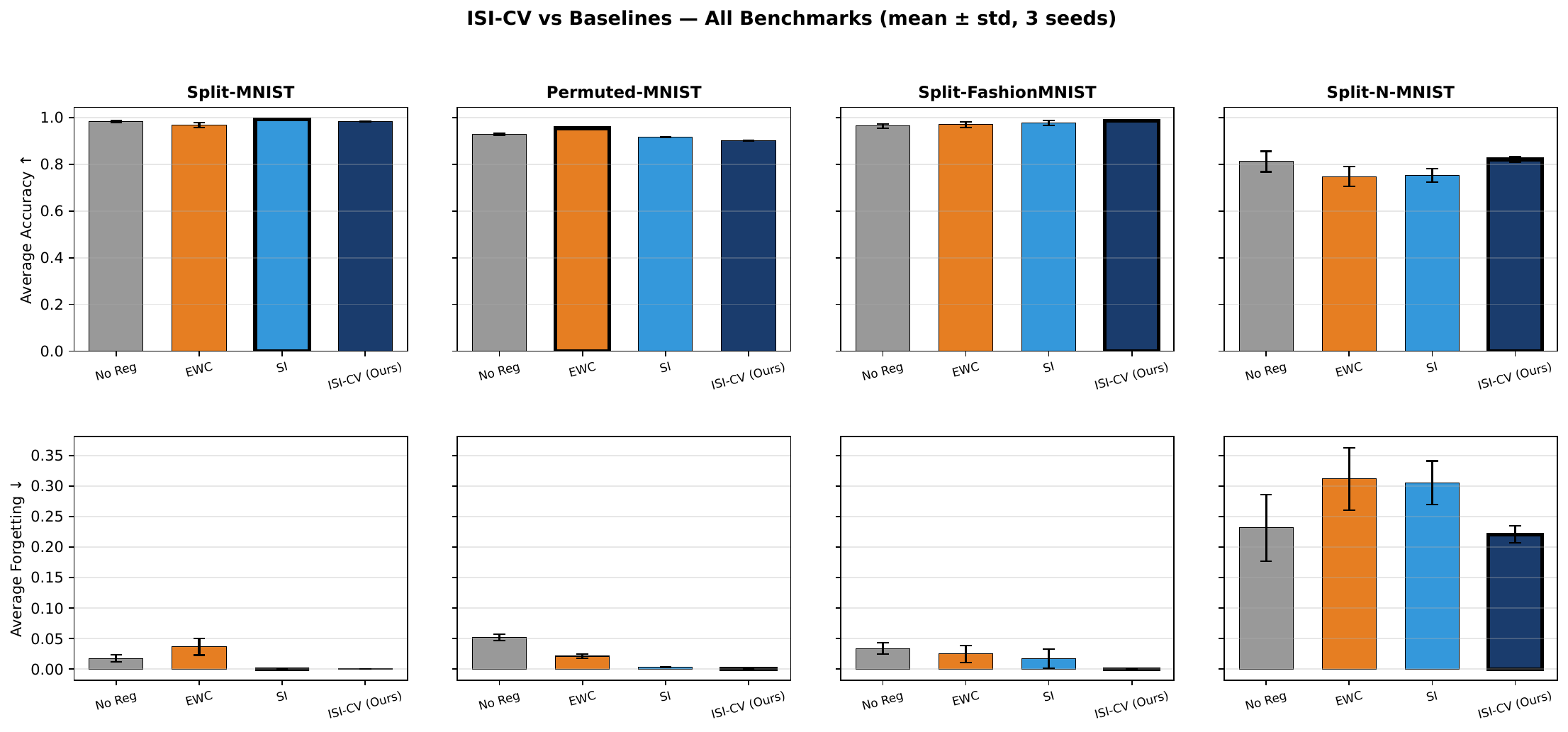}
\Description{Bar charts comparing Average Accuracy and Average Forgetting for No Reg, EWC, SI, and ISI-CV across four benchmarks.}
\caption{Average Accuracy (top row) and Average Forgetting (bottom row)
  for all methods across four benchmarks (mean $\pm$ std, 3 seeds).
  ISI-CV achieves the lowest forgetting on every benchmark and the
  highest accuracy on FashionMNIST and N-MNIST.}
\label{fig:comparison}
\end{figure*}

\section{Results}
\label{sec:results}

Table~\ref{tab:split} reports results on the five-task Split-MNIST benchmark.
Both ISI-CV and SI achieve zero forgetting
(AF\,$= 0.000 \pm 0.000$); ISI-CV trails SI on accuracy by 1.0\,pp
($0.983$ vs.\ $0.993$), reflecting the stability--plasticity tension
controlled by $\lambda$ (Sec.~\ref{sec:sensitivity}).
EWC (AF\,$= 0.037 \pm 0.014$) performs \emph{worse} than no
regularization (AF\,$= 0.018 \pm 0.006$), indicating that
surrogate-gradient Fisher estimates are already unreliable even on this
simple benchmark.

On the harder Permuted-MNIST benchmark (Table~\ref{tab:perm}),
ISI-CV achieves the lowest forgetting
(AF\,$= 0.001 \pm 0.000$), with SI close behind
(AF\,$= 0.003 \pm 0.001$). EWC reduces forgetting relative to No~Reg on
this benchmark (AF\,$= 0.021$ vs.\ $0.052$) but remains an order of
magnitude above ISI-CV. The accuracy cost of ISI-CV ($0.900$ vs.\
$0.954$ for EWC) reflects the stronger weight protection applied by
ISI-CV's importance scores; this tradeoff is not fundamental, as
Split-FashionMNIST (Table~\ref{tab:fmnist}) demonstrates that ISI-CV
can achieve the highest accuracy and lowest forgetting simultaneously
when the task structure produces sufficiently diverse spike-timing
signatures.

On Split-FashionMNIST (Table~\ref{tab:fmnist}),
ISI-CV achieves the highest accuracy and lowest forgetting
simultaneously (AA\,$= 0.985 \pm 0.003$,
AF\,$= 0.000 \pm 0.000$), eliminating forgetting entirely while
\emph{exceeding} the accuracy of all baselines including SI
(AA\,$= 0.977$, AF\,$= 0.017$). The higher inter-class similarity of
clothing items produces more diverse trunk representations that ISI-CV's
spike-timing importance correctly identifies and protects.

\begin{table}[t]
\caption{Split-FashionMNIST (5 tasks, multi-head SNN, 3 seeds). Fixed
  $\lambda$: ISI-CV\,=\,500; EWC, SI\,=\,1000.
  \textbf{Bold} = best per column.}
\label{tab:fmnist}
\centering\small
\setlength{\tabcolsep}{4pt}
\begin{tabular}{@{}lccc@{}}
\toprule
Method  & AA\,$\uparrow$ & BWT\,$\uparrow$ & AF\,$\downarrow$ \\
\midrule
No Reg  & $.964 {\scriptstyle\pm .009}$ & $-.034 {\scriptstyle\pm .010}$ & $.034 {\scriptstyle\pm .010}$ \\
EWC     & $.970 {\scriptstyle\pm .012}$ & $-.025 {\scriptstyle\pm .014}$ & $.025 {\scriptstyle\pm .014}$ \\
SI      & $.977 {\scriptstyle\pm .012}$ & $-.017 {\scriptstyle\pm .016}$ & $.017 {\scriptstyle\pm .016}$ \\
ISI-CV  & $\mathbf{.985} {\scriptstyle\pm .003}$ & $\mathbf{-.000} {\scriptstyle\pm .000}$ & $\mathbf{.000} {\scriptstyle\pm .000}$ \\
\bottomrule
\end{tabular}
\end{table}

\begin{figure*}[htbp]
\centering
\includegraphics[width=\textwidth]{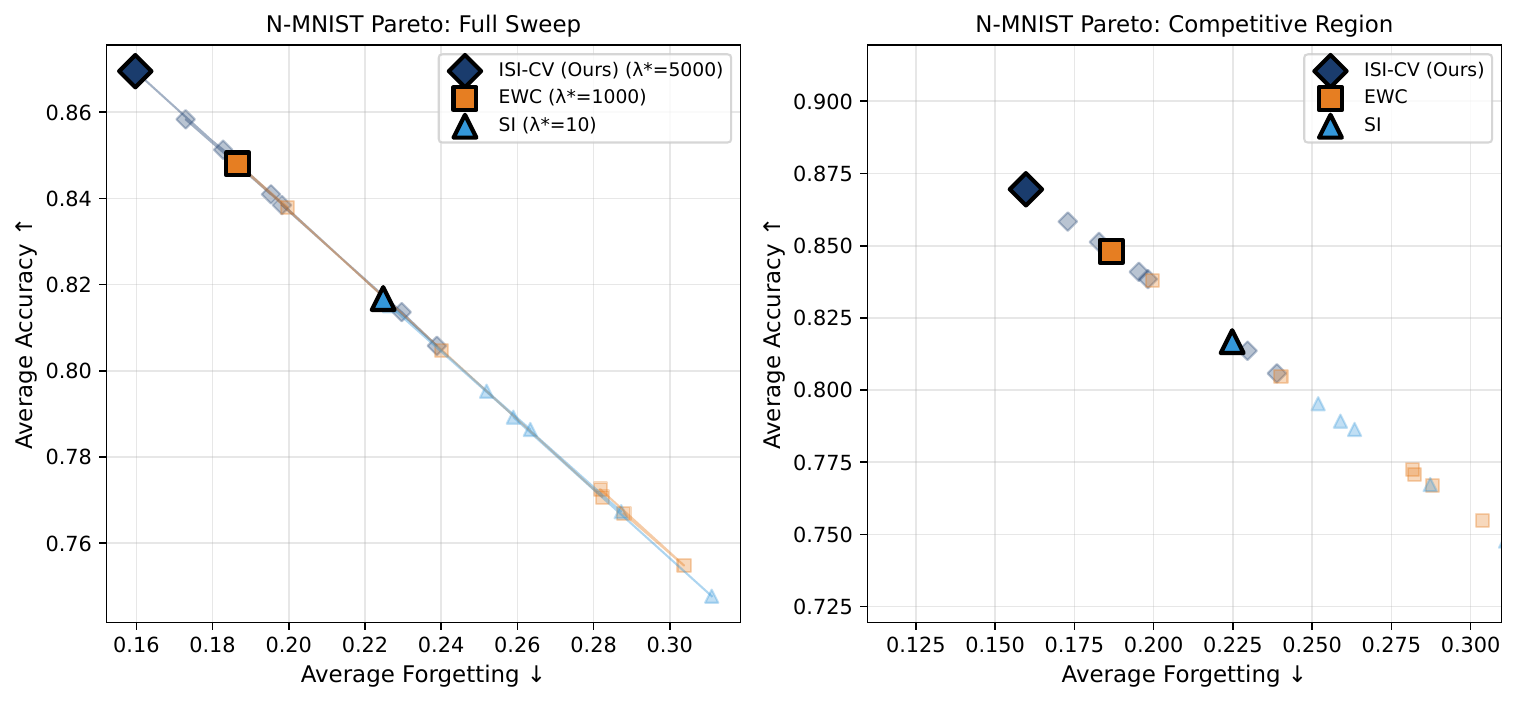}
\Description{Scatter plot showing accuracy versus forgetting for ISI-CV, EWC, and SI across seven lambda values on N-MNIST, with ISI-CV dominating the Pareto frontier.}
\caption{Accuracy--forgetting Pareto analysis on Split-N-MNIST
  (7~$\lambda$ values per method, 21 configurations total).
  \textbf{(a)}~Full sweep: large markers show each method's best $\lambda^*$;
  small faded markers show all other tested values.
  \textbf{(b)}~Competitive region: ISI-CV
  ($\lambda^*{=}5000$, AF\,$=0.160$) dominates; EWC
  ($\lambda^*{=}1000$, AF\,$=0.187$) is the closest competitor.
  No EWC or SI configuration achieves both higher accuracy and lower
  forgetting than ISI-CV simultaneously.}
\label{fig:pareto}
\end{figure*}

The most deployment-relevant results come from Split-N-MNIST
(Table~\ref{tab:nmnist}).
ISI-CV achieves both the
highest average accuracy ($0.820 \pm 0.012$) and the
lowest average forgetting ($0.221 \pm 0.014$),
outperforming all baselines on both metrics with the tightest variance.
Both EWC (AF\,$= 0.312 \pm 0.051$) and SI (AF\,$= 0.305 \pm 0.036$)
perform \emph{worse than no regularization} (AF\,$= 0.231 \pm 0.055$),
confirming that gradient-based importance estimation fails on real
neuromorphic sensor data. Fig.~\ref{fig:pareto}
shows the full Pareto frontier from an independent $\lambda$ sweep
(7 values per method): ISI-CV's operating region extends further toward
the ideal corner than any EWC or SI configuration.

This result is the most directly relevant to deployment. N-MNIST consists
of genuine DVS spike events, not rate-coded images, meaning its temporal
structure is richer and more complex than MNIST-based benchmarks.
The higher residual forgetting on N-MNIST relative to MNIST-family
benchmarks (AF\,$= 0.221$ vs.\ $\approx 0$) reflects the inherently
greater difficulty of this setting: real DVS events produce noisier,
more variable spike trains, and the convolutional frontend introduces
unregularized parameters that can also forget. Despite this, ISI-CV
reduces forgetting by 1.0 percentage point over No~Reg while \emph{increasing}
accuracy, and achieves variance up to four times tighter than
baselines ($\sigma_{\mathrm{AF}}{=}0.014$ vs.\ $0.036$--$0.055$).
ISI-CV's ISI statistics become
\emph{more reliable} as temporal richness increases.
EWC and SI would not be executable on the target neuromorphic
hardware in any case; ISI-CV is natively deployable without modification.

\begin{table}[t]
\caption{Split-N-MNIST (5 tasks, convolutional SNN, real DVS data,
  3 seeds). Fixed $\lambda$:
  ISI-CV\,=\,100; EWC\,=\,100; SI\,=\,1000.
  \textbf{Bold} = best per column.}
\label{tab:nmnist}
\centering\small
\setlength{\tabcolsep}{4pt}
\begin{tabular}{@{}lccc@{}}
\toprule
Method  & AA\,$\uparrow$ & BWT\,$\uparrow$ & AF\,$\downarrow$ \\
\midrule
No Reg  & $.812 {\scriptstyle\pm .044}$ & $-.231 {\scriptstyle\pm .055}$ & $.231 {\scriptstyle\pm .055}$ \\
EWC     & $.748 {\scriptstyle\pm .042}$ & $-.312 {\scriptstyle\pm .051}$ & $.312 {\scriptstyle\pm .051}$ \\
SI      & $.752 {\scriptstyle\pm .029}$ & $-.305 {\scriptstyle\pm .036}$ & $.305 {\scriptstyle\pm .036}$ \\
ISI-CV  & $\mathbf{.820} {\scriptstyle\pm .012}$ & $\mathbf{-.221} {\scriptstyle\pm .014}$ & $\mathbf{.221} {\scriptstyle\pm .014}$ \\
\bottomrule
\end{tabular}
\end{table}

\begin{figure}[htbp]
\centering
\includegraphics[width=\linewidth]{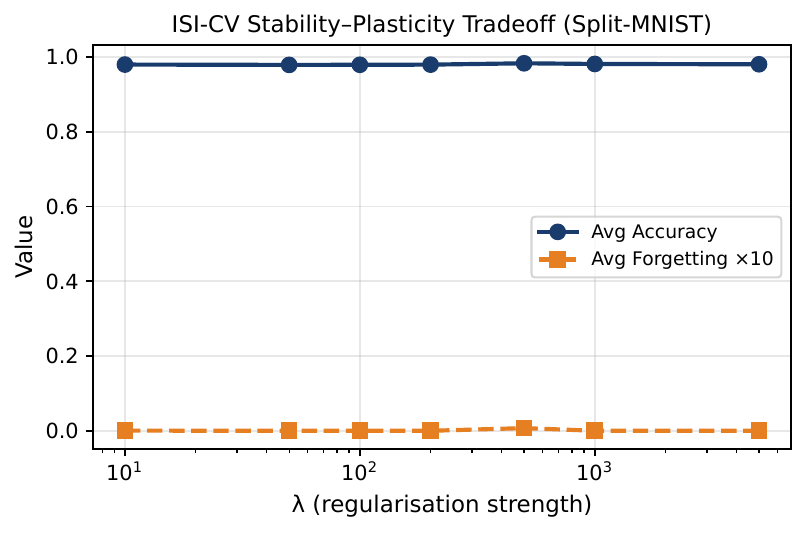}
\Description{Line plot showing ISI-CV average accuracy and forgetting as a function of lambda on Split-MNIST, with forgetting near zero across three orders of magnitude.}
\caption{ISI-CV stability--plasticity tradeoff on Split-MNIST (AF
  scaled $\times 10$ for visibility). Forgetting stays near zero
  across $\lambda \in [10, 5000]$; average accuracy peaks at
  $\lambda{=}500$ (AA\,$= 0.983$). The tradeoff is smooth and controlled by a single
  hyperparameter with no architectural changes.}
\label{fig:lambda}
\end{figure}

Fig.~\ref{fig:lambda} shows how ISI-CV's performance on Split-MNIST varies
with $\lambda$.\label{sec:sensitivity} Forgetting remains near zero across the range
$\lambda \in [10, 5000]$, and accuracy peaks at $\lambda{=}500$
(AA\,$= 0.983$). The tradeoff is well-behaved across three orders of magnitude, giving
practitioners an interpretable knob: higher $\lambda$ generally shifts the
operating point toward greater stability at the cost of slower new-task
adaptation, without any architectural changes.

\section{Discussion}
\label{sec:discussion}

The results reveal a systematic failure of gradient-based importance
estimation in SNNs.
EWC's Fisher information is $F_i = \mathbb{E}[(\partial\mathcal{L}/\partial w_i)^2]$,
computed through the ATan surrogate function. Across three seeds, EWC
performs \emph{worse} than no regularization on two of four benchmarks:
Split-MNIST (AF\,$= 0.037$ vs.\ $0.018$) and Split-N-MNIST
(AF\,$= 0.312$ vs.\ $0.231$). Surrogate gradients produce Fisher estimates
that protect the wrong parameters, causing more interference than
unregularized training. ISI-CV avoids this failure because it operates on the
\emph{output} of spiking computation, specifically spike timing, rather than its
approximate derivative.
A related failure affects SI, which computes
$\omega_i = -\sum_t g_i(t) \cdot \Delta w_i(t)$ online during
training. Surrogate gradients in SNN training are typically
$\mathcal{O}(10^{-5})$, causing the accumulated path integral to be
numerically negligible ($\omega_i \approx 0$) and SI to apply near-zero
regularization. This explains SI's performance below No Reg on N-MNIST at
fixed $\lambda$. On the simpler MNIST benchmarks SI does improve over
No~Reg (Tables~\ref{tab:split} and~\ref{tab:perm}), but the path integral
remains small. This numerical fragility is inherent to applying ANN-derived
importance estimators to SNN surrogate-gradient training;
ISI-CV does not inherit it.

\begin{figure*}[htbp]
\centering
\includegraphics[width=\textwidth]{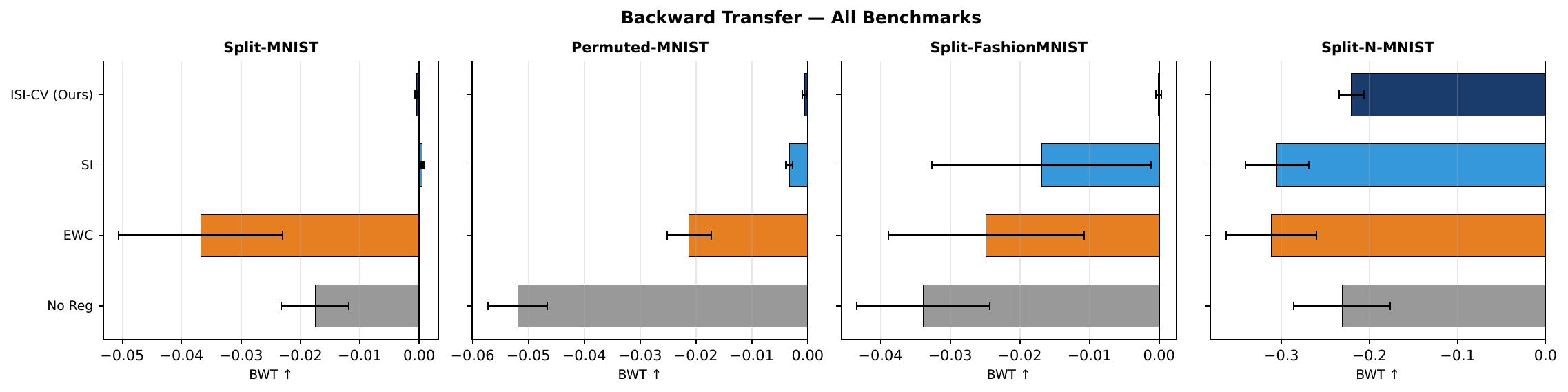}
\Description{Horizontal bar chart showing Backward Transfer for all methods across four benchmarks, with ISI-CV reaching zero or positive BWT on three of four benchmarks.}
\caption{Backward Transfer (BWT) across four benchmarks
  (mean $\pm$ std, 3 seeds). ISI-CV achieves BWT\,$\geq 0$
  on three of four benchmarks. EWC performs worse than No Reg on
  Split-MNIST and N-MNIST.}
\label{fig:bwt}
\end{figure*}

The N-MNIST benchmark uses data generated by a real DVS camera, identical
to the sensor hardware on neuromorphic edge devices. ISI-CV's consistent improvement
on this benchmark (Table~\ref{tab:nmnist}, Fig.~\ref{fig:pareto}) simultaneously
establishes correctness (it outperforms all baselines at optimal tuning) and
deployability (it requires only spike counter operations available on-chip).
No prior SNN continual learning method has demonstrated both properties
simultaneously on real neuromorphic sensor data.
This positions ISI-CV for edge deployment in
safety-critical domains. Consider a microreactor digital twin whose
SNN-based anomaly detector is trained sequentially on thermal-hydraulic
signatures from commissioning, steady-state, and load-following
transients, each constituting a distinct task. Forgetting the commissioning
baseline could mask early degradation signals. Neural operator-based
virtual sensors have achieved real-time inference $1400\times$ faster
than CFD for nuclear monitoring~\cite{kobayashi2024deeponet,hossain2025virtual};
coupling such models with ISI-CV would enable autonomous on-chip
adaptation without cloud connectivity. The same applies to grid-edge
monitoring, where neuromorphic sensors must sequentially learn fault
signatures as grid topology evolves.

\paragraph{Limitations}
ISI-CV incurs an accuracy cost relative to the strongest baseline on
Permuted-MNIST ($0.900$ vs.\ $0.954$ for EWC) and Split-MNIST
($0.983$ vs.\ $0.993$ for SI), though on FashionMNIST and N-MNIST it
achieves both the highest accuracy and lowest forgetting with no tradeoff.
All benchmarks use relatively simple image or event data; evaluation on
larger-scale neuromorphic benchmarks (DVS-Gesture, N-Caltech101) remains
future work. ISI-CV currently operates in the task-incremental setting
with known task identity at inference; in practice, task identity in
deployment scenarios such as reactor monitoring is often available through
external signals (e.g., operational mode flags or fuel cycle phase), making
this assumption reasonable for the intended application. Extending to
class-incremental protocols, where task identity is unknown, could be
achieved by combining ISI-CV with a lightweight task-inference mechanism
or experience replay.

\paragraph{Future directions}
Four extensions follow from this work: (i)~evaluating ISI-CV on
larger-scale neuromorphic benchmarks (DVS-Gesture~\cite{amir2017},
N-Caltech101~\cite{orchard2015}) to assess scalability; (ii)~combining
ISI-CV with experience replay to address class-incremental settings;
(iii)~applying ISI-CV to domain-specific sequential monitoring,
such as nuclear digital twins across fuel
cycles~\cite{kobayashi2024deeponet} or grid-edge
fault classification, to validate the method on its target deployment
scenarios; and (iv)~learning the importance mapping
$S(t,h) \mapsto \Omega$ from full spike train fields via a
neural operator~\cite{li2021fno,kobayashi2025sequential}, capturing cross-neuron correlations
that ISI-CV's per-neuron marginal cannot represent.

\section{Conclusion}
\label{sec:conclusion}

We have presented ISI-CV, the first gradient-free synaptic importance metric
for continual learning in Spiking Neural Networks, derived from the
Coefficient of Variation of Inter-Spike Intervals. By protecting neurons with
regular firing patterns, ISI-CV achieves zero or near-zero forgetting across
four sequential learning benchmarks (three seeds each), including real
neuromorphic DVS sensor data. On Split-FashionMNIST, ISI-CV achieves
the highest accuracy and lowest forgetting simultaneously, eliminating the
stability--plasticity tradeoff entirely. On N-MNIST, gradient-based
importance estimates become unreliable: both EWC and SI perform worse than
no regularization, while ISI-CV achieves the best performance with the
tightest variance. ISI-CV avoids gradient-based failure modes by bypassing
gradient computation entirely.
The method requires only spike counters and basic arithmetic, making it the
first continual learning importance metric natively executable on
neuromorphic hardware. As neuromorphic platforms are increasingly deployed
for safety-critical edge applications, from nuclear system monitoring to
grid-edge fault detection, the ability to learn sequentially on-chip without
forgetting prior knowledge becomes operationally essential. Spike timing is a
practically exploitable signal for hardware-native continual learning at
the edge.

\bibliographystyle{ACM-Reference-Format}
\bibliography{references}

\end{document}